%% file: main.tex
\documentclass[runningheads]{llncs}

\usepackage{graphicx}

\input{tex/00_abbrev_and_macros}

\input{tex/00_abbrev_pgfplots}

\usepackage{tikz}
\usepackage{comment}
\usepackage{array}
\usepackage{booktabs}
\usepackage{multirow}
\usepackage{subcaption}
\usepackage{wrapfig}
\usepackage{amsmath,amssymb} 
\usepackage{color}
\usepackage{listings}
\usepackage{algorithm}

\usepackage[accsupp]{axessibility}  

\usepackage[colorlinks,citecolor=ForestGreen]{hyperref}

\begin{document}
\pagestyle{headings}
\mainmatter
\def\ECCVSubNumber{3760}  

\title{Granularity-aware Adaptation for Image Retrieval over Multiple Tasks}


%
\author{Jon Almaz\'an\inst{1} \and 
 Byungsoo Ko\inst{2} \and \newline
 Geonmo Gu\inst{2} \and 
 Diane Larlus\inst{1} \and
 Yannis Kalantidis\inst{1}}
\authorrunning{J. Almaz\'an et al.}
\institute{$^{1}$NAVER LABS Europe 
$^{2}$NAVER Corp.}

\maketitle

\input{tex/00_abstract}

\input{tex/01_introduction}
\input{tex/02_related}
\input{tex/03_method}
\input{tex/04_experiments}
\input{tex/05_conclusions}

\bibliographystyle{splncs04}
\bibliography{biblio}

\appendix
\input{tex/99_appendix}

\end{document}

%% file: tex/00_abbrev_and_macros.tex
\usepackage{xspace}
\usepackage[dvipsnames, table]{xcolor}

\usepackage{diagbox}

\newcommand{\ykt}[1]{{\color{Brown}{#1}}}

\newcommand{\grappacolor}{Plum}
\newcommand{\grappat}{{\textcolor{\grappacolor}{\texttt{Grappa-T}}}\xspace}
\newcommand{\grappa}{{\textcolor{\grappacolor}{\texttt{Grappa}}}\xspace}
\newcommand{\grappan}{{\textcolor{\grappacolor}{\texttt{Grappa-N}}}\xspace}
\newcommand{\grappaavg}{{\textcolor{\grappacolor}{\texttt{Grappa-avg}}}\xspace}

\newcommand{\teaserfigheight}{4.5cm}

\newcommand{\mrt}{MRT\xspace}
\newcommand{\mrtlong}{Multiple Retrieval Tasks\xspace}

\newcommand{\cub}{CUB\xspace}
\newcommand{\cars}{Cars\xspace}
\newcommand{\sop}{Products\xspace}
\newcommand{\food}{Food-101\xspace}
\newcommand{\aircraft}{Aircraft\xspace}
\newcommand{\flowers}{Flowers\xspace}

\newcommand{\cPone}{\textcolor{red}{$\cP_1$}\xspace}
\newcommand{\cPtwo}{\textcolor{BurntOrange}{$\cP_2$}\xspace}
\newcommand{\cPthree}{\textcolor{Dandelion}{$\cP_3$}\xspace}
\newcommand{\cPfour}{\textcolor{LimeGreen}{$\cP_4$}\xspace}
\newcommand{\cPfive}{\textcolor{OliveGreen}{$\cP_5$}\xspace}
\newcommand{\cPsix}{\textcolor{NavyBlue}{$\cP_6$}\xspace}
\newcommand{\cPseven}{\textcolor{Violet}{$\cP_7$}\xspace}
\newcommand{\cPeight}{\textcolor{Fuchsia}{$\cP_8$}\xspace}
\newcommand{\cPi}{\textcolor{Fuchsia}{$\cP_i$}\xspace}

\usepackage{pifont}
\newcommand{\xmark}{\ding{55}}%
\newcommand{\cmark}{\ding{51}}%

\newcommand{\rp}{$\cR$P\xspace}
\newcommand{\mapatr}{MAP@R\xspace}

\newcommand{\diffup}[1]{{\color{OliveGreen}{($\uparrow$ #1)}}}
\newcommand{\diffdown}[1]{{\color{BrickRed}{($\downarrow$ #1)}}}

\makeatletter
\DeclareRobustCommand\onedot{\futurelet\@let@token\@onedot}
\def\@onedot{\ifx\@let@token.\else.\null\fi\xspace}

\def\eg{\emph{e.g}\onedot} 
\def\ie{\emph{i.e}\onedot}

\def\etal{\emph{et al}\onedot}
\makeatother

\newcommand{\myparagraph}[1]{\vspace{0.1cm}\noindent\textbf{#1.}}

\newcommand{\cA}{\mathcal{A}}

\newcommand{\cC}{\mathcal{C}}
\newcommand{\cD}{\mathcal{D}}

\newcommand{\cF}{\mathcal{F}}

\newcommand{\cL}{\mathcal{L}}
\newcommand{\cM}{\mathcal{M}}
\newcommand{\cN}{\mathcal{N}}
\newcommand{\cO}{\mathcal{O}}
\newcommand{\cP}{\mathcal{P}}

\newcommand{\cR}{\mathcal{R}}

\newcommand{\cT}{\mathcal{T}}

\newcommand{\cZ}{\mathcal{Z}}

\newcommand{\vK}{\mathbf{K}}

\newcommand{\vQ}{\mathbf{Q}}

\newcommand{\vU}{\mathbf{U}}

\newcommand{\vc}{\mathbf{c}}

\newcommand{\vh}{\mathbf{h}}

\newcommand{\vx}{\mathbf{x}}

\newcommand{\vz}{\mathbf{z}}

%% file: tex/00_abbrev_pgfplots.tex
\usepackage{pgfplots, pgfplotstable}
\pgfplotsset{compat=newest}
\usepgfplotslibrary{fillbetween}
\usepackage{tabu}

\newcommand{\supc}{black}
\newcommand{\dinoc}{black}
\newcommand{\pglevelc}{Lavender}
\newcommand{\grappac}{\grappacolor}
\newcommand{\oraclec}{gray}

\newcommand{\leg}[1]{\addlegendentry{#1}}

\pgfmathsetmacro{\stdgrad}{30}
\pgfmathsetmacro{\markersizeteaser}{2pt}

\tikzset{every mark/.append style={solid}}
\pgfplotsset{
	grid=both, width=\linewidth, 
	legend cell align=left, 
	ylabel near ticks,
    xlabel near ticks,
    every tick label/.append style={font=\footnotesize},
}

\pgfplotsset{
    dino/.style={thick, color=\dinoc, mark=*,mark size=\markersizeteaser, only marks},
    sup/.style={thick, color=\supc, mark=x,mark size=\markersizeteaser, only marks},
    grappa/.style={thick, color=\grappac, mark=*,mark size=\markersizeteaser, only marks},
    grappapg/.style={thick, color=\grappac, mark=o,mark size=\markersizeteaser, only marks},
    grappaavg/.style={thick, color=\grappac, mark=diamond,mark size=\markersizeteaser, only marks},
    oracle/.style={thick, color=\oraclec, mark=triangle*,mark size=\markersizeteaser, only marks},
    pglevels/.style={thick, color=\pglevelc, mark=*,mark size=\markersizeteaser, only marks},
    pglevelsline/.style={thick, color=\pglevelc, mark=*,mark size=\markersizeteaser, dotted},
    lineonly/.style={thick, mark size=0pt},
    dinolineonly/.style={mark=none, color=\dinoc},
}

%% file: tex/00_abstract.tex
\begin{abstract}
Strong image search models can be learned for a specific domain, \ie set of labels, provided that some labeled images of that domain are available. A practical visual search model, however, should be versatile enough to solve multiple retrieval tasks simultaneously, even if those cover very different specialized domains. Additionally, it should be able to benefit from even unlabeled images from these various retrieval tasks. This is the more practical scenario that we consider in this paper. We address it with the proposed \grappa, an approach that starts from a strong pretrained model, and adapts it to tackle multiple retrieval tasks concurrently, using only unlabeled images from the different task domains. We extend the pretrained model with multiple independently trained sets of adaptors that use pseudo-label sets of different sizes, effectively mimicking different pseudo-granularities. We reconcile all adaptor sets into a single unified model suited for all retrieval tasks by learning fusion layers that we guide by propagating pseudo-granularity attentions across neighbors in the feature space. Results on a benchmark composed of six heterogeneous retrieval tasks show that the unsupervised \grappa model improves the zero-shot performance of a state-of-the-art self-supervised learning model, and in some places reaches or improves over a task label-aware oracle that selects the most fitting pseudo-granularity per task.
\end{abstract}

%% file: tex/01_introduction.tex
\begin{wrapfigure}[19]{R}{0.45\linewidth}
  \begin{center}
    \resizebox{\linewidth}{!}{
      \input{plots/teaser_small}
    }
  \end{center}
\caption{\textbf{\grappa} is an unsupervised method that trains \textit{a single model} with higher zero-shot performance (measured with $\cR$-Precision or \rp)
than the pretrained DINO~\cite{caron2021dino} model, over several retrieval tasks.
}
\label{fig:teaser}
\end{wrapfigure}
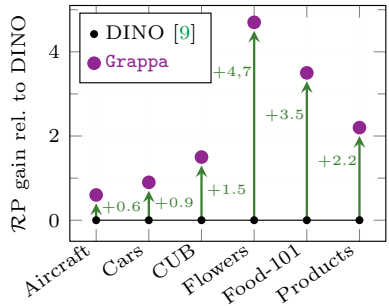

\section{Introduction}
\label{sec:introduction}

The last few years have witnessed progress on image retrieval: successful models can be trained, provided that a set of labeled images from the domain of interest (not necessary from the same categories) is available for training, as in the common deep metric learning scenario.
Those models are as powerful as they are specialized: it has been shown, and we confirm in our experiments, that one model carefully tailored for one domain (\eg bird species) tend to perform poorly to a neighboring yet different domain (\eg dog breeds).

Here, we argue that a practical visual search system should be able to solve multiple retrieval tasks simultaneously, without needing to explicitly specialize for each task.
Consider for example a visual search system specialized to fauna and flora. In such \ykt{a} system, the image database covers a broad range of fine-grained domains, \eg from searching among different insect species to different kinds of mushrooms. For the system to also handle coral species, it should be as simple as providing a set of unlabeled coral images.

In parallel, the field has worked towards pretraining large and generic models for visual representations that can be used, often as a black box, to extract features for new tasks. Among those, models trained in a self-supervised way have shown to be versatile to various target tasks, including image retrieval~\cite{grill2020byol,caron2021dino}.

In this work, we assume access to such a large pretrained model that already provides good zero-shot performance. We also assume access to an unlabeled set of images possibly from multiple tasks. We propose to adapt the initial model so it performs even better on multiple image retrieval tasks \textit{simultaneously}, \ie when this
same adapted model is used to extract features for all tasks.

This raises two questions.
First, \textit{how should we perform adaptation?} Fine-tuning is prohibitively costly especially for large pretrained models, and does not always transfer well.
As an alternative to fine-tuning, and inspired by an early work on multi-task training~\cite{rebuffi17learning} and a recent trend in natural language processing~\cite{houlsby2019adapter,pfeiffer21adapter}, we propose to use adaptor layers.
Adaptors are embedded in between architecture blocks, and are the only weights learned, the ones from the original pretrained model remaining fixed. Our experiments show that this composite architecture allows for a versatile adaptation of a strong initial model by adjusting only a small percentage of the model parameters. 

Second, \textit{how should we reconcile various retrieval tasks in a single model?} 
A retrieval task focuses on a given set of visual concepts, often associated to a particular granularity. Yet, unlike in classification for which the granularity is known beforehand, the granularity of a retrieval task is context dependent, and depends on the gallery of images where visual search is performed. 
We therefore propose learning different sets of adaptors, each set tailored to one specific granularity. 
As we assume that training images are unlabeled, not even to indicate the retrieval task they correspond to, we propose to automatically define levels of granularity by partitioning the training set into more and more clusters. As a result, each partition corresponds to a different set of pseudo-labels. We then independently train one set of adaptors for each pseudo-granularity.

Next, we need to reconcile these different sets of adaptors into a single multi-purpose retrieval model. One option is to combine them with a naive fusion mechanism. 
The resulting model improves results on all retrieval tasks, showing the clear benefit of a multi-granularity understanding of the data.
Another option is to go one step further and to achieve adaptor fusion via attention propagation. 
In this case, we require consistency between the adaptor attention of nearest neighbors in the feature space. We observe this fusion mechanism further improves the model.

To summarize, our contribution is threefold.
First, we palliate the absence of image and task labels by creating sets of pseudo-labels, with the goal of approximating any possible granularities in a given set of retrieval tasks.
Second, we propose a way to extend transformer-based architectures with adaptors, and a training framework that tailors individual sets of adaptors to different pseudo-granularities.
Third, we propose a number of ways for fusing the adapter features, \eg via augmentation invariance or via propagating attention from neighbors in the image features space.
We validate our approach on a collection of datasets for deep metric learning and we show that \grappa improves over the successful DINO pretrained model, a model known to already obtain strong zero-shot performance on all these retrieval tasks (see Fig.~\ref{fig:teaser}).

%% file: plots/teaser_small.tex
\begin{tikzpicture}
\begin{axis}[%
  width=\linewidth,
  xtick = {1,2,3,4,5,6},
  xticklabels = {\aircraft,\cars,\cub,\flowers,\food,\sop},
  x tick label style={rotate=35,anchor=east},
  ylabel style={font=\scriptsize}, 
  ylabel = \rp gain rel. to DINO,
    legend pos=north west,
  tick label style={font=\scriptsize},
  legend style={font=\scriptsize}, 
  grid style={gray!1}, 
  ymin=-0.5, ymax=5.1,
  minor y tick num=1,
  height=\teaserfigheight,
  ]

\pgfplotstableread{
d	dino grappa
1   0  0.6
2   0  0.9
3   0  1.5
4   0  4.7
5   0  3.5
6   0  2.2
}{\map}
    \addplot[dino, mark size=\markersizeteaser-1] table[x=d,  y=dino]   \map; \leg{DINO~\cite{caron2021dino}}
    \addplot[grappa,  mark=*, color=\grappacolor] table[x=d,  y=grappa]   \map; \leg{\grappa}
    \addplot[dinolineonly] table[x=d,  y=dino]   \map;

\draw[thick,OliveGreen,->,>=stealth] (axis cs:1,0) -- (axis cs:1,0.4)
    node [anchor=west,OliveGreen] at (axis cs:0.9,0.3) {\tiny{+0.6}};

\draw[thick,OliveGreen,->,>=stealth] (axis cs:2,0) -- (axis cs:2,0.7)
    node [anchor=west,OliveGreen] at (axis cs:1.9,0.4) {\tiny{+0.9}};
    
\draw[thick,OliveGreen,->,>=stealth] (axis cs:3,0) -- (axis cs:3,1.3)
    node [anchor=west,OliveGreen] at (axis cs:2.9,0.7) {\tiny{+1.5}};

\draw[thick,OliveGreen,->,>=stealth] (axis cs:4,0) -- (axis cs:4,4.5)
    node [anchor=west,OliveGreen] at (axis cs:3,3.5) {\tiny{+4,7}};

\draw[thick,OliveGreen,->,>=stealth] (axis cs:5,0) -- (axis cs:5,3.3)
    node [anchor=west,OliveGreen] at (axis cs:4,2.5) {\tiny{+3.5}};

\draw[thick,OliveGreen,->,>=stealth] (axis cs:6,0) -- (axis cs:6,2)
    node [anchor=west,OliveGreen] at (axis cs:5,1.4) {\tiny{+2.2}};
    
\end{axis}
\end{tikzpicture}

%% file: tex/02_related.tex
\section{Related Work}
\label{sec:related}

The
task we tackle in this paper
strongly relates to deep metric learning. It requires specific architectural changes of neural networks to extend them with adaptors. Note that our task can be seen as solving a zero-shot problem, 
\ie it requires no labeled data from the downstream datasets and learns a single model for all tasks, 
something fairly uncommon in transfer learning.

\myparagraph{Deep metric learning (DML)}
DML aims to learn a metric between data points that reflects the semantic similarity between them. 
It plays an important role in a wide range of tasks such as image clustering~\cite{hershey2016deep,caron2018deep}, unsupervised learning~\cite{chen2020simple,he2020momentum,caron2020swav}, and visual search~\cite{cao2020unifying,dir,apgem}. 
Recent DML approaches typically learn visual similarity using either a pair-based loss~\cite{chopra2005learning,sohn2016improved,oh2016deep,gu2020symmetrical,ko20embedding} which considers pair-wise similarities, a proxy-based loss~\cite{wang2017normface,wang2018additive,deng2019arcface,gu2020proxy,ko2021learning}, which considers the similarity between samples and class representative proxies, or a contextual classification loss~\cite{zhai2019normsoftmax,boudiaf2020mutual,elezi2020group,seidenschwarz2021intrabatch}.
In most cases, DML approaches finetune an ImageNet pretrained model for \emph{each} target retrieval task, and each of those finetuned models fall short when applied to other retrieval tasks.
We aim at a more versatile visual search system that handles multiple retrieval tasks with a \emph{single} model.

\myparagraph{Neural architectures with adaptation layers} Adaptation layers (or adaptors) have emerged \cite{rebuffi17learning,houlsby2019adapter,pfeiffer21adapter,wang2021k} as a way to avoid common problems rising in sequential finetuning or multi-task learning when trying to finetune large pretrained models to solve multiple tasks, namely the issues of catastrophic forgetting~\cite{mccloskey89catastrophic} and task imbalance.
Rebuffi \etal~\cite{rebuffi17learning} were the first to introduce adaptors 
to visual recognition tasks, adapting a convolutional model to many classification tasks.
Adaptors have also been used with transformer architectures for natural language processing~\cite{houlsby2019adapter}; bottleneck layers are added to all the blocks of a pretrained model and finetuned, keeping the underlying model fixed.

Recently, Pfeiffer \etal~\cite{pfeiffer21adapter} introduced a way to share knowledge between adaptors using an adaptor fusion layer within a two-stage learning framework: adaptors are trained independently in the first stage, they are kept fixed while only the fusion layer is trained in the second stage.
All the methods mentioned above still result in models that specialize to a single task; \eg~\cite{pfeiffer21adapter} learns a separate fusion layer per downstream task, whereas we would like to learn a single model for all tasks.

\myparagraph{Zero-shot problems}
The field has recently taken an interest in pretraining large models, sometimes called zero-shot models, using large quantities of data. Those have been shown to be versatile and applicable to many target tasks. Among them, self-supervised models~\cite{chen2020simclr,caron2020swav,he2020momentum,zbontar2021barlow,caron2021dino} are trained using self-defined pseudo-labels as supervision and typically millions of images (\eg from ImageNet~\cite{deng2009imagenet}).
Recent works~\cite{goyal2021self,tian2021divide} exploit even larger yet uncurated, sets of unlabeled images to enhance the quality of the learned representations. 
Others~\cite{radford2021clip,jia2021scaling,zhai2022lit} have leveraged multiple modalities, \eg training visual representations so they are similar to the textual representations of their associated text. 
Those self-supervised or multimodal methods offer excellent initialization to be finetuned for a wide range of downstream tasks. 
Sometimes they are used in a zero-shot setting: a single model is used as a feature extractor, typically to solve multiple tasks. This is the regime we study here,
but we further assume that a small amount of unlabeled data from the downstream tasks exists.

\myparagraph{Relation to other transfer tasks}
The idea of transferring a model trained for a given task to a related one has become central to computer vision~\cite{razavian2014cnn,sariyildiz2021cog}, and appears in many research fields such as task transfer~\cite{zamir2018taskonomy}, domain adaptation~\cite{csurka2017domain} or self-supervised learning~\cite{gidaris2018rotnet,noroozi2016jigsaw,caron2021dino}. 
Yet, in all those, the initial model is only a starting point and it is typically not only extended, 
but also retrained for each task of interest, leading to a multitude of specialized models. In our work, we need a \textit{single} model to perform well across retrieval tasks. In that regard, this work is closer to zero-shot transfer of the large pretrained models discussed above.
Also related are Mixtures of Experts (MoE)~\cite{yuksel2012experts,shazeer2017outrageously,puigcerver2021scalable,riquelme2021scaling}, an ensembling technique that decomposes a predictive problem into subtasks, training one expert for each. Although MoE architectures may look similar to ours at first glance, they typically rely on gating and pooling mechanisms that learn to predict, in a supervised way, which experts to trust, and how to combine them. Similar to typical transfer approaches, they build one specialized model for each target task. Here, we focus on a purely unsupervised task: no labels are provided to indicate image semantic content nor the retrieval task images belong to.

%% file: tex/03_method.tex
\section{A granularity-aware multi-purpose retrieval model}
\label{sec:method}

In this section we present \grappa, a method for adapting a pretrained model to multiple retrieval tasks simultaneously, in an unsupervised way.
We first formalize our task, \ie visual search over several retrieval tasks using a single model (Sec.~\ref{sec:background}). We then present an overview of the approach (Sec.~\ref{sec:overview}). Next, we detail each step, \ie building multiple granularities (Sec.~\ref{sec:clustering}), learning adaptors using granularity-aware pseudo-labels (Sec.~\ref{sec:adaptors}), and learning to fuse them by propagating adaptor attention across feature space neighbors (Sec.~\ref{sec:fusion}).

\subsection{Background}
\label{sec:background}

Our task of interest, \textbf{visual search on multiple retrieval tasks}, can be seen as a variant of the standard deep metric learning (DML) task. The most common protocol in DML is to a) split the \textit{classes}%
\footnote{We will use the term \textit{classes} to refer to sets of images with the same label, whether the latter represents object instances or fine-grained classes.} 
into disjoint train and test sets of labels; b) learn a separate model for each retrieval task on the corresponding train split; c) perform retrieval on all images of the (unseen) test split of classes.

Our setting has several key differences. First, we solve multiple retrieval tasks \textit{simultaneously}. This means that we do not learn one model for each but a \textit{single} model that will be used for all tasks. Second, we only assume access to a set of \textit{unlabeled} images from each retrieval task, and do not have access to labeled training sets, unlike standard DML methods. Even more challenging, unlabeled training images are provided jointly without knowing which target retrieval task they correspond to nor the total number of retrieval tasks.

More formally, let $\cT$ be the set of $m$ retrieval tasks that we want to simultaneously tackle. Each task $\cT^t$ is associated with a training and a test set. At train time, we are provided with a fused training set $\cD$ composed of the union of all training sets of the $m$ datasets in $\cT$. 
As mentioned earlier, images are not associated to any class or task label. 

With so many unknowns on the target retrieval tasks, an obvious choice is to \textbf{start with a large pretrained model}.
Self-\cite{chen2020simclr,he2020momentum,caron2021dino} or weakly-\cite{radford2021clip,jia2021scaling} supervised learning have been shown to lead to strong models that generalize well and exhibit high zero-shot transfer performance.
We assume that we are given such a model. 
Here, we base our work on the recently proposed Visual Transformer (ViT)~\cite{dosovitskiy2021an}, a popular, efficient, and highly performing architecture, pretrained in a self-supervised way with DINO~\cite{caron2021dino}.

We set our pretrained model $\cM$ to be a ViT with $L$ transformer layers and an input patch size $P \times P$ pixels. 
Input image $\vx \in \mathbb{R}^{H\times W \times C}$ is reshaped into a sequence of $T$ flattened 2D patches where $T=HW/P^2$. The transformer uses constant latent vector size $D$ through all of its layers, so flattened patches are first mapped to $D$ dimensions with a trainable linear projection and concatenated in~$\vh^0$, together with a prepended learnable \texttt{[class]} token and added position embeddings. The transformer encoder~\cite{Vaswani2017AttentionIA} consists of alternating blocks of multi-headed self-attention (MSA) and MLP (which contain two layers with a GELU non-linearity). LayerNorm (LN) is applied before every block, and residual connections after every block. Formally, each layer of $\cM$ (shown with a gray background on Fig.~\ref{fig:architecture}, left) is given by: 

\begin{equation}
    \vh^l = \text{MLP}(\text{LN}(\tilde{\vh}^l)) + \tilde{\vh}^l, \quad \tilde{\vh}^l = \text{MSA}(\text{LN}(\vh^{l-1})) + \vh^{l-1},
\end{equation}
for $l = \{1\ldots L\}$. The image representation $\vz$ is the output of the \texttt{[class]} token after the last layer $\vh^L$, \ie $\vz = \text{LN}(\vh^{L})[$\texttt{class}$]$. We refer the reader to~\cite{dosovitskiy2021an} for more details about the ViT architecture.

\begin{figure}[t]
    \centering
    \resizebox{\textwidth}{!}{
    \begin{subfigure}[c]{0.18\textwidth}
        \centering
        \includegraphics[width=\textwidth]{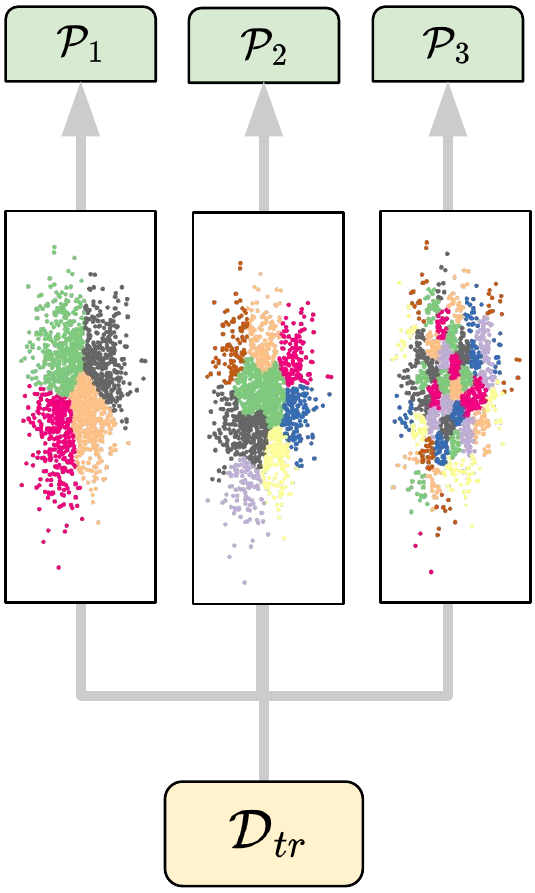}
        \caption{\scriptsize{Step 1}}
        \label{fig:training_pseudo}
    \end{subfigure}
    \begin{subfigure}[c]{0.63\linewidth}
        \centering
        \includegraphics[width=\textwidth]{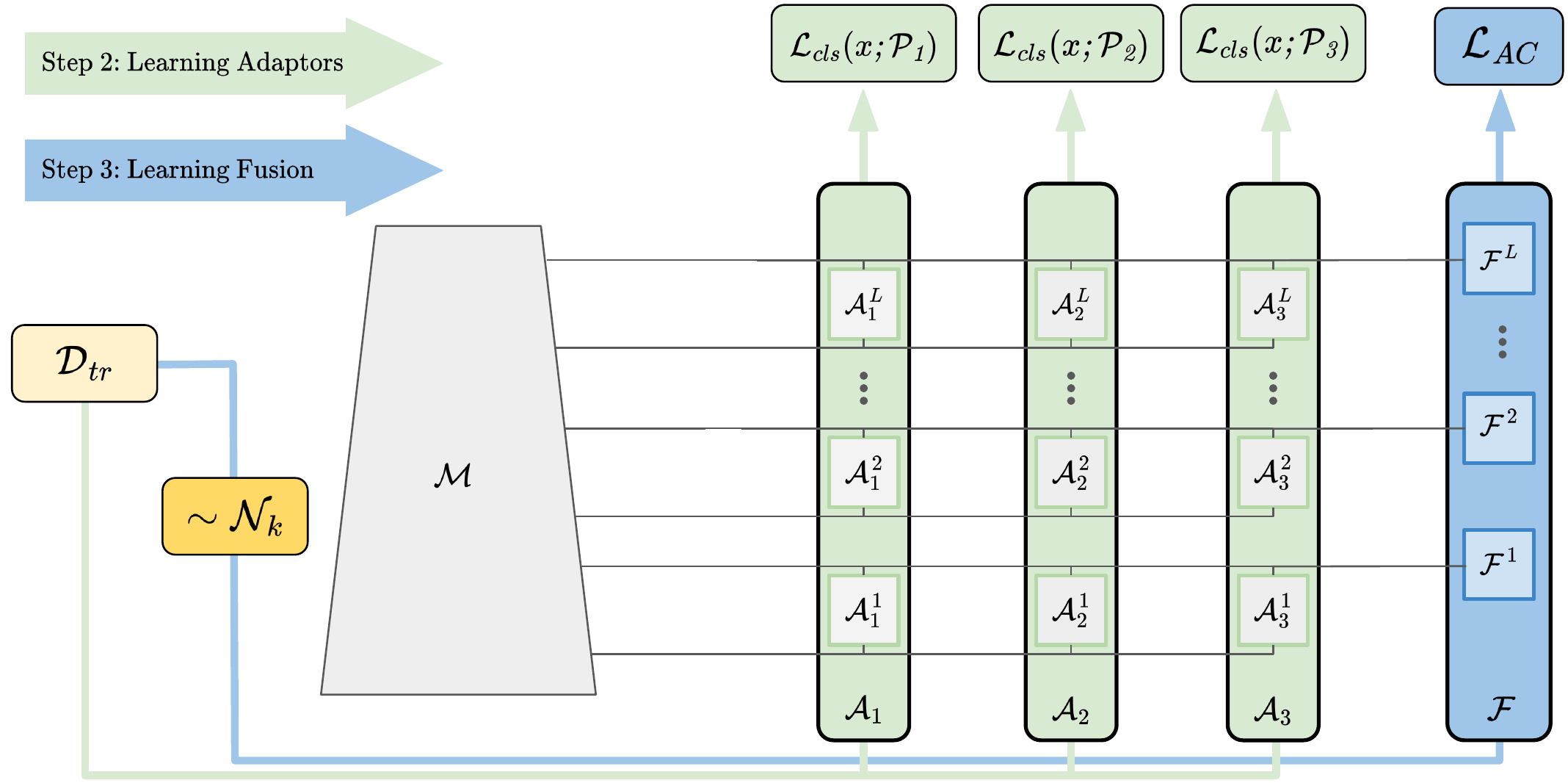}
        \caption{{\scriptsize{\textcolor{YellowGreen}{Step 2}} and {\textcolor{RoyalBlue}{Step 3}}}}
        \label{fig:training_stages}
    \end{subfigure}%
}
\caption{\textbf{Training of the proposed \grappa}. Left: granularities correspond to pseudo-labels $\cP_i$ obtained by multiple clusterings of the feature space (Step 1). Right: we learn the granularity-aware adaptors (Step 2, in green), and then learn how to fuse them (Step 3, in blue). For this example, $N$=3. }
\label{fig:training}
\end{figure}

\subsection{Method overview}
\label{sec:overview}

Our method builds on the VIT~\cite{dosovitskiy2021an} model $\cM$ pretrained with DINO~\cite{caron2021dino}, that we treat as an architectural backbone. We extend and train it in an unsupervised way using $\cD$.
The training process consists of three steps (summarized in Fig.~\ref{fig:training}):
\begin{itemize}
    \item[$\bullet$] \textbf{Step 1: Learning pseudo-labels.} We build multiple sets of pseudo-labels. Each set partitions the feature space using clustering and corresponds to a different pseudo-granularity. 
        This process is illustrated in Fig.~\ref{fig:training_pseudo}. 
    \item[$\bullet$] \textbf{Step 2: Learning adaptors for each pseudo-label set independently.} We learn a set of adaptors specific to each pseudo-granularity using a classification loss. This process is depicted by the green arrows in Fig.~\ref{fig:training_stages}. 
    \item[$\bullet$] \textbf{Step 3: Learning to fuse adaptors.} We learn a set of fusion layers to merge the outputs of multiple adaptors using a transformation invariance or an attention propagation loss, \ie neighboring images should have similar attentions over adaptors. This process is depicted by the blue arrows in Fig.~\ref{fig:training_stages}. 
\end{itemize}

These three stages lead to a single model we denote as $\cM^*$, that unifies the multiple granularities, and consists of:
the pretrained model $\cM$ used as a \textit{frozen} backbone (its parameters are kept fixed during the entirety of the process), embedded adaptors $\cA_i$, and fusion layers $\cF$. 
This single model is used as a unique feature extractor for \textit{all} retrieval tasks considered in our benchmark. We denote our method \textbf{\grappa}, that stands for learning \textcolor{\grappacolor}{\textbf{Gr}}anularity-aware \textcolor{\grappacolor}{\textbf{A}}daptors by Attention \textcolor{\grappacolor}{\textbf{P}}ro\textcolor{\grappacolor}{\textbf{pa}}gation. 
The following subsections detail the learning stages.

\subsection{Step 1: Learning pseudo-labels}
\label{sec:clustering}

We would like to build multiple sets of pseudo-labels such that they 
partition the feature space at different `granularities'. We can approximate this partitioning by estimating multiple sets of clusters while varying the number of centers. 

In practice, we extract features using the pretrained model $\cM$. Let $\vz = f(\vx; \cM)$ be the feature of an image $\vx \in \cD$. Let the set of all features for training set $\cD$ be $\cZ = \{f(\vx; \cM), \forall \vx \in \cD \}$. 
To get multiple sets of pseudo-labels, we cluster the full set of features $\cZ$ into sets of centroids $\cC_i, i = 1..N$, of respectively $k_i$ clusters, where $k_i$ gets monotonically larger as $i$ approaches $N$.
This produces $N$ sets of pseudo-labels $\cP_1, \ldots, \cP_N$. 
For each pseudo-label set $\cP_i$, an image $\vx \in \cD$, is associated to a pseudo-label given by $\cP_i(\vx) = \arg \min_{\vc \in \cC_i} ||\vz - \vc||$, for $\vz = f(\vx; \cM)$. 
We rely on the vanilla $k$-means clustering algorithm~\cite{kmeans} with $k$-means++~\cite{arthur2006k} initialization, a common choice for the size of our benchmark.
For even larger datasets, more scalable variants could be used, like hierarchical~\cite{philbin2007object}, approximate~\cite{avrithis2012approximate}, or quantized~\cite{avrithis2015web} $k$-means.%
Note that other works have used k-means to define pseudolabels~\cite{caron2018deep,yan20clusterfit}. Yet, our work is the first to learn multiple sets and subsequently use all of them.

\subsection{Step 2: Learning adaptors for each pseudo-label set}
\label{sec:adaptors}

Given the $N$ sets of pseudo-labels computed in the previous step, we now would like to learn adaptors tailored to each pseudo-label set, \ie to each pseudo-granularity. 
We use the pretrained model $\cM$ as a backbone and extend it by embedding an adaptor at every layer. We then learn the adaptor parameters while keeping the backbone ones frozen. We learn a set of $L$ adaptors for \textit{each} pseudo-granularity in an independent way.

\begin{figure}[t]
    \centering
    \includegraphics[width=\linewidth]{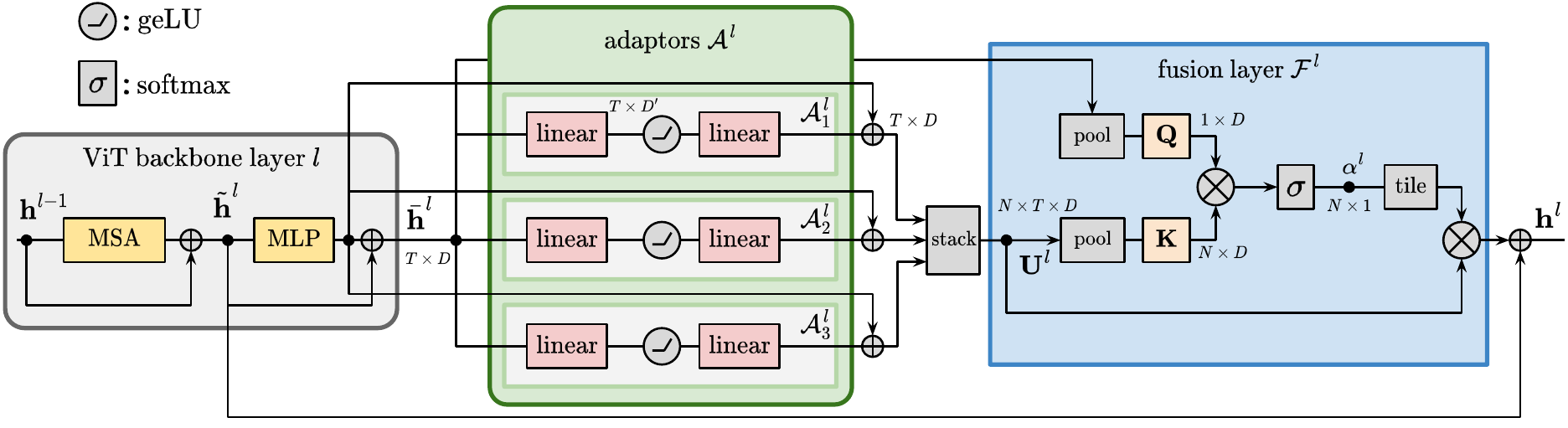}
    \caption{\textbf{Architecture of the $l^\text{th}$ layer of the \grappa model}, for $N$=3 adaptors.}
    \label{fig:architecture}
\end{figure}

\myparagraph{Adaptor architecture}
Recent works in natural language processing \cite{houlsby2019adapter,philip2020monolingual,wang2021k} have embedded adaptor layers in transformer architectures. We follow a similar design and embed $L$ adaptors, one at the end of each transformer layer of $\cM$.
 
Formally, we learn a separate set of adaptors $\cA_i$ for each pseudo-label set $\cP_i$, $i = \{1,\ldots,N\}$. Each set $\cA_i$ consists of $L$ adaptors, denoted as $\cA_i^1,\ldots, \cA_i^L$. These adaptors are bottleneck layers with an intermediate dimensionality of $D^\prime$ (where $D^\prime < D$), a GELU layer~\cite{hendrycks2016gaussian} in between, and a residual connection at the end. 
Since we are modifying the architecture of $\cM$ by interleaving it with blocks, we need to revisit notations. The output of layer $l$ in $\cM$ (after the basic ViT block) is now defined as $\bar{\vh}^l = \text{MLP}(\text{LN}(\tilde{\vh}^l)) + \tilde{\vh}^l$. The output of the new layer $l$ (the original VIT block combined with an adaptor) can still be denoted as $\vh^l$.
Details of the overall architecture are shown in Fig.~\ref{fig:architecture}.

\myparagraph{Learning the adaptors}
Given a set of pseudo-labels $\cP_i$, we can learn the parameters of the set of adaptors $\cA_i$ via a supervised cross entropy loss. Specifically, we use the norm-softmax loss~\cite{wang2018additive,wang2017normface,liu2017sphereface,zhai2019normsoftmax} that, for image $\vx$, is given by:
\begin{equation}
    \cL_{cls}(\vx; y) = -\log \frac{\exp(\gamma \cos \theta_y)}{\sum_{y^\prime = 1}^{k_i} \exp(\gamma \cos \theta_{y^\prime}) }, 
    \label{eq:normsoftmax}
\end{equation}
where $\gamma$ is a scale factor, $\cos \theta_y$ is the cosine similarity to the classifier of class $y$, and the loss is guided by the pseudo-labels, \ie $y = \cP_i(\vx)$. After learning the parameters for each set $\cP_i$, we keep the adaptors and discard the 
classifiers.

\subsection{Step 3: Learning to fuse adaptors}
\label{sec:fusion}

The process described in Sec.~\ref{sec:adaptors} leads to $N$ separate sets of adaptors, each tailored to a different pseudo-granularity.
The next step is to unify all adaptor sets into a single architecture. To that end, we append (\ie stack) the $N$ adaptors for each layer \textit{in parallel}, as shown  in Fig.~\ref{fig:architecture}. We then concatenate adaptor outputs in a tensor $\vU^l \in \mathbb{R}^{N\times T \times D}$ for each layer $l = \{1,\ldots,L\}$, where each row corresponds to the output of one adaptor for this layer. Here, another residual connection is added, giving the model the opportunity to bypass the adapter if needed.  
Tensor $\vU^l$ is therefore given by $\vU^l =  \{ \cA_i^l(\bar{\vh}^l) + MLP(LN(\tilde{\vh}^l)), i=1..N\}$
and is then fed, together with $\bar{\vh}^l$, to a fusion layer, as detailed below. 

\myparagraph{First Option: Fusion by average pooling}
A straightforward way of fusing the outputs of the $N$ adaptors is to treat them as equally important and average them. The fusion layer therefore is simply an average pooling layer that takes tensor $\vU^l \in \mathbb{R}^{N\times T \times D}$ as input and computes the mean over its first dimension.
We refer to this simpler version of our approach as \grappa-avg.

\myparagraph{Second option: Learning to fuse}
Treating all adaptors as equally important for any input image goes against our intuition that different retrieval tasks are more related to certain granularities, and hence more suited for the corresponding adaptors.
We therefore design a fusion layer with trainable parameters, that can learn to weigh the different adaptor outputs.
We use a simple dot-product self-attention architecture over the sequence of $N$ adaptor outputs. Yet, we make two crucial modifications to the vanilla \textit{query-key-value} self-attention: a) To learn an \textit{image-level} attention, we average over the $T$ spatial tokens; 
b) Given that we want to fuse the adaptors but do not want to alter the adaptor representations, we omit the linear projection of the value branch, and only learn projections for the query and key branches that affect the re-weighting of adaptor features.

Specifically, the fusion layer learns an attention vector of size $N$ over the adaptors, given inputs $\bar{\vh}^l$ and $\vU^l$, by $\cF^l(\bar{\vh}^l, \vU^l) = \alpha^l(\bar{\vh}^l, \vU^l) \vU^l$, where vector $\alpha^l(\bar{\vh}^l, \vU^l) \in \mathbb{R}^N$ is given by:

\begin{equation}
\alpha^l(\bar{\vh}^l, \vU^l) = \text{softmax}\left(\frac{\left(\vQ\sum_T\bar{\vh}^l\right) \left(\vK\sum_T\vU^l\right)^{T}}{\sqrt{D}}\right)
\end{equation}
where $l = \{1,\ldots,L\}$, $\vQ$ and $\vK$ are linear projections of size $D \times D$. A final residual connection is added after the fusion layer. The architecture details of a complete layer are shown in Fig.~\ref{fig:architecture}. The latter comprises the ViT block, the adaptors, and the fusion layer, all appended in a residual fashion.

Given pretrained model $\cM$ and multiple sets of adaptors, one way to build a single model is to select one set of adaptors per image. 
This amounts to guessing which pseudo-granularity best fits each image.
We argue that, in a generic visual search system, ``picking a granularity'' for a query image depends less on the image content than on the retrieval task, \ie the gallery used at test time.
Given a dog image query, for example, the only way to know if we are looking for any dog image or only images of the same dog breed, is by looking at the local structure of the gallery around that image.
Both scenarios might favor different representations; our system reconciles them by learning a combination of adaptors. 
Obviously, we do \textit{not} have access to the test images during training. Yet, we assume access to unlabeled set of images $\cD$, representative of the target retrieval tasks, or at least of their granularity. Again, these images are provided without \textit{task labels}, we do not know which retrieval task they correspond to. 

Without any other supervisory signal, we argue that the local neighborhood in the feature space of the training set $\cD$ can be used to approximate the ``granularity'' of a query.
In other words, we assume that visually similar images from $\cD$ should yield similar attention vectors over the sets of adaptors. We therefore propose to learn to fuse adaptors using a loss on neighboring image pairs in the feature space.
In this step, the backbone model $\cM$ and the adaptors remain frozen. 
The fusion layer only learns $\vK$ and $\vQ$, two linear projections that are multiplied to give the attention vectors $\alpha^l$, for each ViT encoder $l$. 
This means that any loss applied to this fusion step \textit{only re-weights} adaptor features. 
We denote the final model, composed of the backbone with all embedded adaptors and their fusion, as $\cM^*$, and the corresponding feature extractor as $f^*(\vx, \cM^*)$.

\myparagraph{Attention propagation loss} 
As mentioned, we propose to train the fusion layer leveraging the assumption that neighboring image pairs in the feature space should use similar attentions over adaptors.
Let $\cN_k(\vx;\cD)$ denote the $k$ nearest neighbors of $\vx$ from dataset $\cD$. We define  neighbors $(\vx_i, \vx_j)$ as a pair of inputs such that $\vx_j \in \cN_k(\vx_i;\cD)$. 
Although neighbors could be built using the pretrained model $\vz = f(\vx, \cM)$ (static $k$-NN), the representations $\tilde{\vz} = f^*(\vx,\cM^*)$ from the learned model $\cM^*$ 
provide a better estimation. 
This requires to  periodically update neighbors during training (in practice we do it at every epoch).
Given a pair of neighboring features, we bring their adaptor attentions close to each other and strive for \textit{attention consistency}.

Attention consistency is enforced using the pairwise Barlow Twins loss~\cite{zbontar2021barlow}.
Given a batch of image pairs, the loss is defined over the output representations $\Tilde{\vz_i} = f^*(\vx_i; \cM^*), \Tilde{\vz_j} = f^*(\vx_j; \cM^*)$ of our model, computed over the $D\times D$ cross-correlation matrix $C$ and averaged over the batch, \ie:
\begin{equation}
    \cL_{BT} =\sum_n (1 - C^{nn})^2 + \beta \sum_n \sum_{m \neq n} (C^{nm})^2,  C_{nm} = \frac{\sum_b g(\hat{\vz_i})^{b,n} g(\hat{\vz_j})^{b,m}}{\sqrt{\sum_b( g(\hat{\vz_i})^{b,n})^2} \sqrt{\sum_b(g(\hat{\vz_j})^{b,m})^2}},
    \label{eq:bt}
\end{equation}
where $b$ iterates over pairs in the batch, $n$ and $m$ iterate over feature dimensions, $\beta$ is a hyperparameter and $g(\cdot)$ is a MLP projector appended to the model and discarded after training. We refer the reader to~\cite{zbontar2021barlow} for more details. 

Originally, \ie in~\cite{zbontar2021barlow}, this loss was defined over two transformed versions of the same image ($\vx_i = t(\vx), \vx_j = t(\vx)$). 
When image pairs are created using image transformations, Eq.(\ref{eq:bt}) defines a transformation consistency (TC) loss or $\cL_{TC}$.
This is a variant that we consider in our benchmark, referred to as \grappat. 
However, we are interested in applying this loss on neighboring pairs in the feature space, $(\vx_i, \vx_j)$ such as $\vx_j \in \cN_k(\vx_i;\cD)$, and using it for attention propagation. In this case, we depart from \cite{zbontar2021barlow} and follow the recent TLDR method~\cite{kalantidis2022tldr} which uses the Barlow Twins loss over neighbor pairs for learning a feature encoder for dimensionality reduction.
Similarly, we use the Barlow Twins loss on image pairs defined using the k-NN graph.
We denote the loss in Eq.(\ref{eq:bt}) as an \textit{attention consistency} (AC) loss, or $\cL_{AC}$, and refer to this variant as \grappan. 

%% file: tex/04_experiments.tex
\section{Experiments}
\label{sec:experiments}
In this section we validate the proposed \grappa on several retrieval tasks. These tasks are collected in a new benchmark that we introduce, called \mrt. It unifies 6  fine-grained classification datasets under a retrieval setting. We show statistics and present the evaluation protocol of this benchmark in Sec.~\ref{sec:mrt}, we present the methods we compare in Sec.~\ref{sec:methods} and we report all results in Sec.~\ref{sec:results}.

\myparagraph{Implementation details} We use ViT-Small \cite{dosovitskiy2021an} as a backbone architecture, with a patch size of 16 pixels and DINO \cite{caron2021dino} pre-trained weights. We generate $N$=8 sets of pseudo-labels on the training set of \mrt,
respectively composed of 256, 1024, 4096, 8,192, 16,384, 32,768, 65,536, and 131,072 clusters.
We learn a set of adaptors for each pseudo-label set, 
using the norm-softmax loss from Eq.(\ref{eq:normsoftmax}) and an Adam optimizer with a learning rate and weight decay of 0.001.
We train the fusion layer over these adaptors using the Barlow Twins loss \cite{zbontar2021barlow} and the LARS optimizer~\cite{you2018lars}. 
We used the same hyper-parameters for the scaling and $\beta$ as suggested in \cite{zbontar2021barlow}, and a learning rate and weight decay of 0.5 and 0.001.

\myparagraph{Evaluation metrics}
Recent works~\cite{musgrave2020metric,fehervari2019unbiased} in DML have questioned standard evaluation metrics (\ie Recall@1) and argue they are not fair. Thus, we report the R-Precision (\rp) and \mapatr metrics recently introduced in~\cite{musgrave2020metric}.

\subsection{\mrtlong (\mrt) Benchmark}
\label{sec:mrt}

\myparagraph{Data} The \mrtlong (\mrt) benchmark combines the 6 following fine-grained datasets under a retrieval setting: \aircraft~\cite{maji2013aircraft}, \cars~\cite{krause2013cars}, \cub~\cite{wah2011cub}, \flowers~\cite{nilsback2008flowers}, \food~\cite{bossard2014food}, and Stanford online products (\sop)~\cite{oh2016deep}. We follow standard practice in the DML community and, for each dataset, assign the first half of the classes (ordered alphabetically) for training and the second half for testing. We then combine images from all the training splits into a single training set $\cD$ of 133,339 images and discard their class and dataset labels. This is the training set we use to learn the pseudo-labels, as well as the adaptor and the fusion parameters. We show statistics for all datasets in Table~\ref{tab:datasets}.

\input{tables/tab_datasets}

\myparagraph{Evaluation protocol} 
Models are trained on the combined training set $\cD$ without task nor class labels. Evaluation is performed on the test split of each task independently, following a leave-one-out protocol: each image is used as a query once to rank all the other images in the test set.
For evaluation, relevance is defined according to class labels and we report mean average precision (\mapatr or mAP) and $\cR$-Precision (\rp) over all queries.

\subsection{Compared methods}
\label{sec:methods}

\myparagraph{Baselines}
First and foremost, we compare with the DINO pretrained visual transformer
of~\cite{caron2021dino}, a self-supervised model trained on ImageNet1K.
It obtains impressive zero-shot performance on retrieval and constitutes a very strong baseline.\footnote{We chose DINO over the CLIP~\cite{radford2021clip} model as the training set of CLIP is not public.
Therefore the data in \mrt might be part of its 400M image-text training pairs.} 
We denote this baseline as \textbf{DINO} or simply $\cM$ in Table~\ref{tab:results} and Fig.~\ref{fig:mrt}.

\input{tables/tab_results_two_metrics}

The \grappa architecture adds an extra set of parameters in the form of adaptors and fusion layers. To verify that improvements do not simply come from these extra parameters, we report results with a second baseline that has the same number of parameters as the \grappa models, but uses no pseudo-granularity-based adaptors nor the proposed attention consistency loss. Instead of following Step 2, we randomly initialize adaptors and finetune them when learning the fusion. For the latter, we use the Barlow Twins loss from Eq.(\ref{eq:bt}), with transformation consistency, and train it on the training set of \mrt, similar to \grappa. We denote this baseline: \textbf{$\cM^*$ (random)}. 

\myparagraph{Proposed}
We report results for models with adaptor fusion
described in Sec~\ref{sec:method} together with results for individual adaptors.  
More precisely, we build $N$ pseudo-label sets on the training set of \mrt and train $N$ sets of adaptors on these pseudo-labels independently. We report their individual performance as \textbf{$\cM$ + \cPi}. As mentioned above, we use DINO as a backbone and keep it frozen. 

Then, using \mrt's training set again, we combine these $N$ adaptors into a single model and train the fusion layer using i) a Barlow Twins loss on the final representation when creating pairs from two augmented views, reported as \grappat, and ii) our proposed Attention Consistency framework which relies on the local neighborhood, reported as \grappan. 
We also report results for the case where the fusion is an average pooling layer, denoted as \grappaavg.

\myparagraph{An adaptor selector oracle}
What if we could choose the best performing pseudo-granularity for each retrieval task? Obviously, this requires access 
to the test set labels. This also results in a different representation per task, which departs from the universal representation we seek to learn. 
For these reasons, we only consider this variant as an oracle, and its results should not be compared with others. 
We still provide it as a reference, showing how much could be achieved if we set the attention as 
a one-hot vector that only enables the best possible pseudo-granularity adaptor.
We denote the oracle as \textbf{$\cO$}.

\subsection{Results}
\label{sec:results}

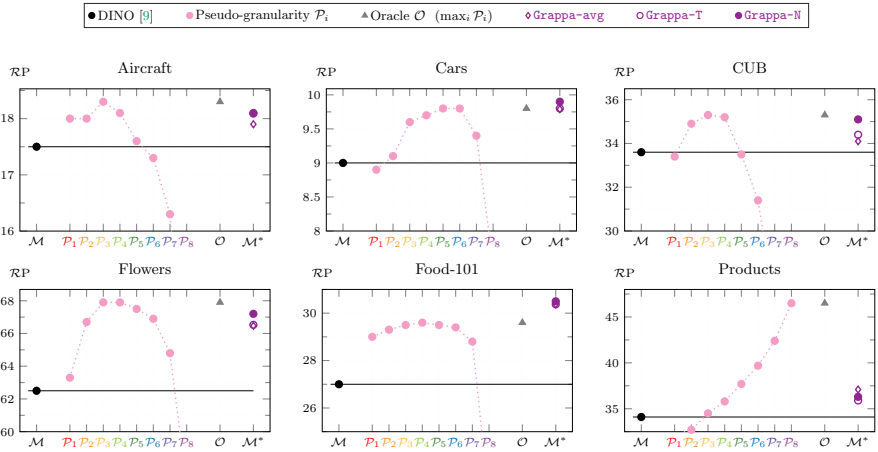
\begin{figure}[t]
    \centering
    \input{plots/per_ds_legend}
    \resizebox{\textwidth}{!}{
    \begin{tabular}{m{.5\textwidth}m{.5\textwidth}m{.5\textwidth}}
        \input{plots/per_ds_aircraft} &
        \input{plots/per_ds_cars}&
        \input{plots/per_ds_cub} \\
        \input{plots/per_ds_flowers} &
        \input{plots/per_ds_food} &
        \input{plots/per_ds_sop} \\
    \end{tabular}
} 
\caption{\textbf{Results per dataset in \mrt.} All 6 datasets use the same model.}
\label{fig:mrt}
\end{figure}

We present our results
in Table~\ref{tab:results} and Fig.~\ref{fig:mrt}. 
Again, note that, unlike the common DML experimental setting, we use \textit{a unique model for all retrieval datasets} and \textit{no class nor task labels during training}. We make the following observations.

\myparagraph{Baselines} 
We confirm the initial observation~\cite{caron2021dino} that, for common DML datasets, \emph{DINO is a very strong baseline}. 
It achieves good performance even on the more challenging metrics \rp and \mapatr. 
Also, it turned out to be very challenging to improve over DINO by keeping its backbone frozen and embedding extra modules.
We did our best to learn the additional modules from scratch, but were unsuccessful. Rows 2-3 of Table~\ref{tab:results} report the best results after hyper-parameter tuning for a single set (no fusion) and for 8 sets of adaptors with fusion. Embedding randomly initialized modules typically deteriorates the performance.
This makes separately trained pseudo-granularity adaptors all the more important.

\myparagraph{Using a single set of adaptors}
In rows 4-12 of Table~\ref{tab:results} we report results when only using adaptors from a single pseudo-granularity; these results are also visualized as lavender-colored points in Fig.~\ref{fig:mrt}. We observe that, for each dataset, there exists at least one pseudo-label set that improves over DINO, and that the best one (reported as Oracle $\cO$) is different for different retrieval tasks.
The oracle results use separate models, and selecting the best one for each task requires access to labels.
Considering each set of adaptors as a separate model, some improve over DINO on several retrieval tasks, showing that even individual pseudo-granularities-specific sets of adaptors can be useful. Yet, as we will see next, higher gains can be achieved by fusing multiple adaptors into a single model.

\myparagraph{Fusion and attention consistency}
The bottom part of Table~\ref{tab:results} shows the performance of the final model $\cM^*$ that combines adaptors from all eight pseudo-granularities (\cPone $\ldots$ \cPeight).
We see that using the proposed attention consistency (AC) loss over neighborhood pairs results in an even stronger model that improves over DINO and over the \grappat variant that uses pairs from different augmentations of the same image, in all datasets. 
In fact, we see that in most cases (\ie apart from \sop), the proposed AC loss is able to give results \textit{on-par with the oracle $\cO$} that would select the right pseudo-granularity for each dataset. Moreover, for \food and \cars, \grappa outperforms the best adaptor for the dataset, showing that this oracle is not necessarily an upper bound: combining pseudo-granularities is beneficial.
It is also worth noting that the simpler and parameter-free fusion of the \grappaavg model is still improving over DINO in all cases, while it is also the best performing \grappa variant on the instance-level Products dataset.

\begin{figure}[t!]
\centering
    \resizebox{.8\textwidth}{!}{
    \begin{tabular}{m{0.5in} m{4.8in} }
        DINO & \includegraphics[width=0.99\textwidth]{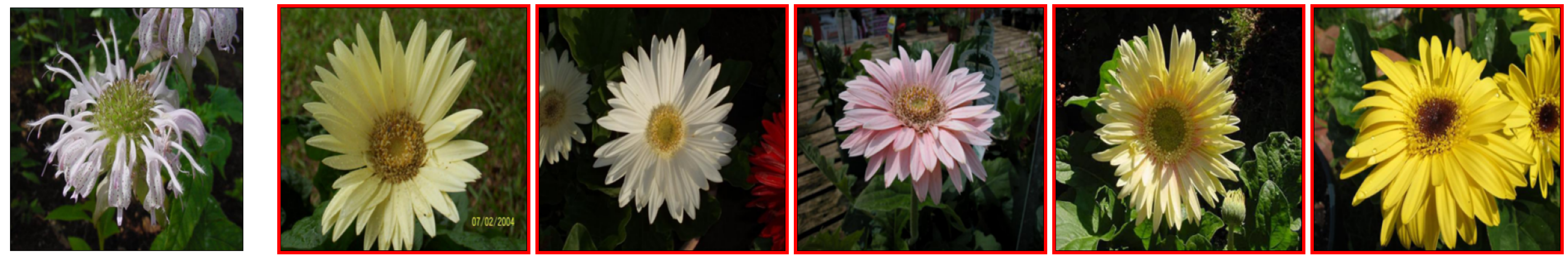}\\ 
        \grappa & \includegraphics[width=0.99\textwidth]{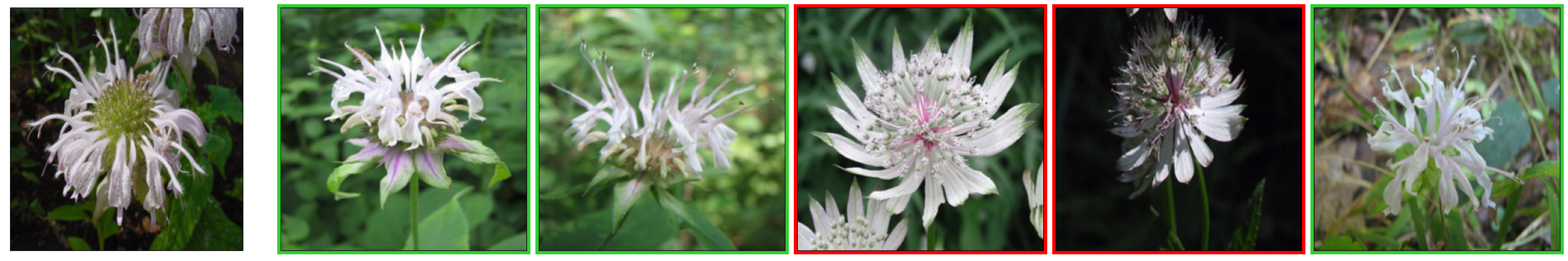} \\
        DINO &\includegraphics[width=0.99\textwidth]{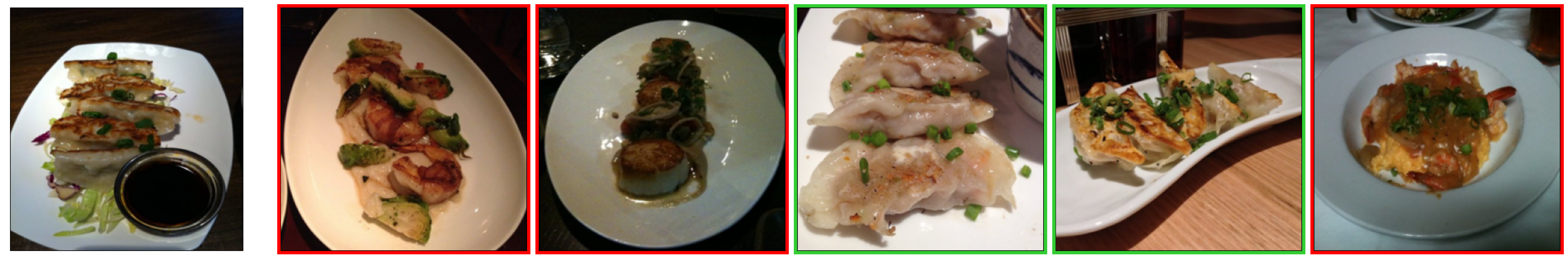} \\ 
        \grappa & \includegraphics[width=0.99\textwidth]{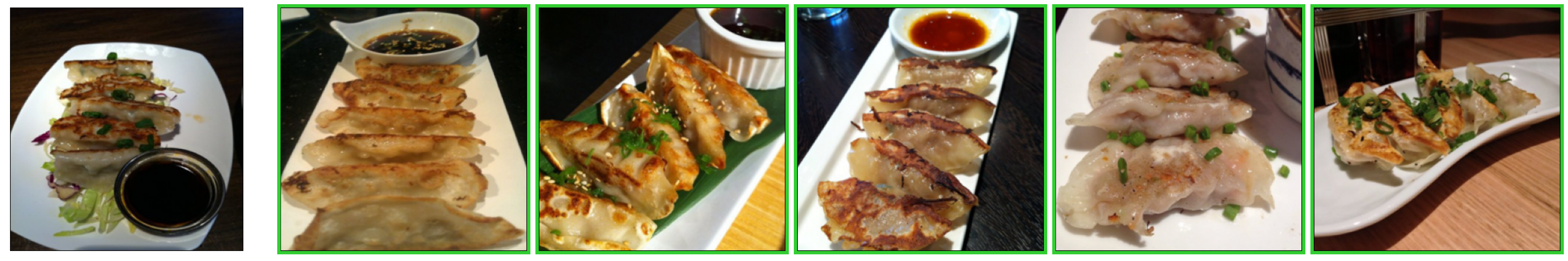} \\
    \end{tabular}
    }
\caption{\textbf{Qualitative results.}
Queries (first column) from \flowers and \food and their top 5 retrieved results (columns 2-6) by DINO~\cite{caron2021dino} and \grappan.}
\label{fig:qualitative}
\end{figure}

\myparagraph{Qualitative results}
Fig.~\ref{fig:qualitative} presents qualitative results for two queries from the \mrt benchmark. We present two cases where \grappan achieves significantly higher recall than DINO, \ie cases where the local feature space is adapted in a way that images from the correct class are closer together.

\myparagraph{Failure cases} 
As we see from Table~\ref{tab:results} and Fig.~\ref{fig:mrt}, 
our fusion mechanism is not able to learn that \cPeight is the best pseudo-granularity for \sop. We attribute this to the different topology of that dataset and the fact that hyperparameters were chosen to optimize performance for the union of the six benchmarks.

%% file: tables/tab_datasets.tex
  \begin{table}[t!]
    \centering
    \caption{\textbf{Statistics of the \mrtlong (\mrt) benchmark.} It is composed of 6 datasets. Classes in train and test are \textit{disjoint}. We provide the number of classes as a reference, but labels are \textit{never} used during training.
    \label{tab:datasets}}
    \setlength{\tabcolsep}{20pt}

    \resizebox{.95\linewidth}{!}{
    \begin{tabular}{l  rr  rr}
    \toprule
         \multirow{2}{*}{Dataset} & \multicolumn{2}{c}{Training} & \multicolumn{2}{c}{Testing}  \\ \cmidrule(l){2-5}
         & \# Classes & \# Images & \# Classes & \# Images \\
    \midrule
    \aircraft~\cite{maji2013aircraft} & 50 & 5,000 & 50 & 5,000 \\    
    \cars~\cite{krause2013cars} & 98 & 8,054 & 98 & 8,131 \\
    \cub~\cite{wah2011cub} & 100 & 5,864 & 100 & 5,924  \\
    \flowers~\cite{nilsback2008flowers} & 51 & 3,870 & 51 & 4,319 \\
    \food~\cite{bossard2014food} & 51 & 51,000 & 50 & 50,000 \\ 
    \sop~\cite{oh2016deep} & 11,318 & 59,551 & 11,316 & 60,502  \\ \midrule
      
    \textbf{\mrt} & 11,668 & 133,339 & 11,665 & 133,876 \\
    \bottomrule
    \end{tabular}
    }

\end{table}

%% file: tables/tab_results_two_metrics.tex
\begin{table}[t]
    \centering
    \caption{\looseness=-1
    \textbf{Results on the Multiple Retrieval Task (\mrt) benchmark.}  We report \mapatr (mAP) and \rp on the six datasets of \mrt, obtained by a single model from those listed in Sec.~\ref{sec:methods}, all unsupervised. The oracle (in gray) is not comparable as it selects the set of adaptors that performs best for each task.}
    \label{tab:results}
    \setlength{\tabcolsep}{5pt}
    \resizebox{\linewidth}{!}{
    \begin{tabu}{l   rr rr rr rr rr rr}
    \toprule
     & \multicolumn{2}{c}{~~~~~~\aircraft} & \multicolumn{2}{c}{~~~~~~\cars} & \multicolumn{2}{c}{~~~~~~\cub} & \multicolumn{2}{c}{~~~~~~\flowers} & \multicolumn{2}{c}{~~~~~~\food} & \multicolumn{2}{c}{~~~~~~\sop} \\
            & ~~~~~~~~\rp         & mAP      & ~~~~~~~~\rp           & mAP        & ~~~~~~~~\rp         & mAP      & ~~~~~~~~\rp           & mAP        & ~~~~~~~~\rp           & mAP         & ~~~~~~~~\rp           & mAP        \\
        
        \midrule

        \textbf{DINO ($\cM$)} & 17.5 & 9.2 & 9.0 & 3.5 & 33.6 & 22.9 & 62.5 & 57.0 & 27.0 & 16.1 & 34.1 & 31.5 \\ \midrule
        \multicolumn{13}{l}{~\emph{Results without pseudo-granularity adaptors}} \\
        $\cM + \cA_1$  & 15.4 & 7.8 & 7.9 & 2.7 & 16.2 & 7.8 & 62.8 & 56.9 & 22.1 & 11.6 & 32.0 & 29.5 \\
        $\cM^*$ (random) & 11.8 & 5.0 & 6.2 & 1.7 & 11.0 & 4.5 & 62.7 & 56.2 & 21.9 & 11.2 & 28.9 & 26.5 \\ \midrule 

        \multicolumn{13}{l}{~\emph{Results using only a single set of pseudo-granularity adaptors}} \\ 
$\cM$ + \cPone & 		18.0 & 9.5 & 8.9 & 3.4 & 33.3 & 22.7 & 63.3 & 57.6 & 29.0 & 17.9 & 32.2 & 29.6 \\
$\cM$ + \cPtwo & 		18.0 & 9.5 & 9.1 & 3.6 & 34.9 & 24.1 & 66.7 & 61.3 & 29.3 & 18.1 & 32.7 & 30.1 \\
$\cM$ + \cPthree & 		18.3 & 9.7 & 9.6 & 3.8 & 35.3 & 24.5 & 67.9 & 62.7 & 29.5 & 18.4 & 34.5 & 31.9 \\
$\cM$ + \cPfour & 		18.1 & 9.5 & 9.7 & 3.9 & 35.2 & 24.3 & 67.9 & 62.6 & 29.6 & 18.4 & 35.8 & 33.1 \\
$\cM$ + \cPfive & 		17.6 & 9.1 & 9.8 & 3.9 & 33.5 & 22.6 & 67.5 & 62.2 & 29.5 & 18.4 & 37.7 & 34.9 \\
$\cM$ + \cPsix & 		17.3 & 8.7 & 9.8 & 3.9 & 31.4 & 20.3 & 66.9 & 61.5 & 29.4 & 18.3 & 39.7 & 36.8 \\
$\cM$ + \cPseven & 		16.3 & 8.0 & 9.4 & 3.7 & 26.5 & 15.8 & 64.9 & 59.1 & 28.8 & 17.8 & 42.4 & 39.6 \\
$\cM$ + \cPeight & 		12.6 & 5.4 & 7.5 & 2.7 & 14.8 & 6.6 & 56.7 & 49.8 & 22.1 & 11.9 & 46.5 & 43.7 \\ \cmidrule(l){2-13}
\rowfont{\color{\oraclec}}
 {Oracle ($\cO$)} &	18.3 & 9.7 & 9.8 & 3.9 & 35.3 & 24.5 & 67.9 & 62.7 & 29.6 & 18.4 & 46.5 & 43.7 \\
 \textcolor{gray}{\ie only}   &	\multicolumn{2}{c}{~~~~~~~~~~\cPthree} &			\multicolumn{2}{c}{~~~~~~~~~~\cPfive--\cPsix} & \multicolumn{2}{c}{~~~~~~~~~~\cPthree} & \multicolumn{2}{c}{~~~~~~~~~~\cPthree--\cPfour} & \multicolumn{2}{c}{~~~~~~~~~~\cPfive} & \multicolumn{2}{c}{~~~~~~~~~~\cPeight} \\ \midrule
    \multicolumn{13}{l}{~\emph{Results using 8 pseudo-granularity adaptors and fusion ($\cM^*$)}} \\ 
\grappaavg
& 17.9 & 9.3 & 9.8 & 3.9 & 34.1 & 23.1 & 66.5 & 61.2 & 30.4 & 19.2  & \textbf{37.1} & \textbf{34.4} \\
\grappat & \textbf{18.1} & \textbf{9.5} & 9.8 & 3.9 & 34.4 & 23.5 & 66.5 & 61.2 & 30.4 & 19.2 & 35.9 & 33.2 \\
\grappan  & \textbf{18.1} & \textbf{9.5} & \textbf{9.9} & \textbf{4.0} & \textbf{35.1} & \textbf{24.1} & \textbf{67.2} & \textbf{61.9} & \textbf{30.5} & \textbf{19.3} & 36.3 & 33.6 \\
~~\emph{vs. DINO} & \diffup{0.6} & \diffup{0.3} & \diffup{0.9} & \diffup{0.5} & \diffup{1.5} & \diffup{1.2} & \diffup{4.7} & \diffup{4.9} & \diffup{3.5} & \diffup{3.2} & \diffup{2.2} & \diffup{2.1} \\
    \bottomrule
    \end{tabu}
    }
    
\end{table}

%% file: plots/per_ds_legend.tex
    \begin{center}
    \resizebox{.8\linewidth}{!}
    {
    \begin{tikzpicture}
        \begin{axis}[%
        hide axis,
        xmin=10, xmax=50,
        ymin=0,ymax=0.4,
        legend columns=-1,
        legend style={/tikz/every even column/.append style={column sep=20pt}},
        legend style={draw=white!15!black,legend cell align=left}
        ]
        \addlegendimage{dino, only marks} \addlegendentry{DINO~\cite{caron2021dino}};
        \addlegendimage{pglevels} \addlegendentry{Pseudo-granularity $\cP_i$};
        \addlegendimage{oracle} \addlegendentry{Oracle $\cO$ ~($\max_i{\cP_i}$)};
        \addlegendimage{grappaavg} \addlegendentry{\grappaavg};
        \addlegendimage{grappapg} \addlegendentry{\grappat};
        \addlegendimage{grappa} \addlegendentry{\grappan};
    
        \end{axis}
    \end{tikzpicture}
    }
    \end{center}

%% file: plots/per_ds_aircraft.tex
\begin{tikzpicture}
\begin{axis}[%
  width=1.1\linewidth,
  title=\aircraft,
  xtick = {-1,1,2,3,4,5,6,7,8,10,12},
  xticklabels = {$\cM$, \cPone,\cPtwo,\cPthree,\cPfour,\cPfive,\cPsix,\cPseven,\cPeight, $\cO$, $\cM^*$},
  ylabel = \rp,
  ylabel style={}, 
  every axis y label/.style={
    at={(ticklabel* cs:1.05)},
    anchor=south,
    font=\scriptsize},
  legend pos=south west,
  tick label style={font=\scriptsize},
  ymin=16,
  xmin=-2, xmax=13,
  minor y tick num=1,
  grid style={gray!1}, 
  height=4.5cm,
]
    \addplot[dino, mark=*, only marks] coordinates {(-1, 17.5)}; 
    \addplot[dinolineonly] coordinates {(-1.5, 17.5) (13, 17.5)};
    \addplot[pglevelsline] coordinates {
(1,	18.0)
(2,	18.0)
(3,	18.3)
(4,	18.1)
(5,	17.6)
(6,	17.3)
(7,	16.3)
(8,	12.6)
}; 
    \addplot[oracle, mark=triangle*] coordinates {(10, 18.3)}; 
    \addplot[grappaavg] coordinates {(12,  17.9)}; 
    \addplot[grappapg] coordinates {(12,  18.09)}; 
    \addplot[grappa, mark=*] coordinates {(12, 18.1)}; 

\end{axis}
\end{tikzpicture}

%% file: plots/per_ds_cars.tex
\begin{tikzpicture}
\begin{axis}[%
  width=1.1\linewidth,
  title=\cars,
  xtick = {-1,1,2,3,4,5,6,7,8,10,12},
  xticklabels = {$\cM$, \cPone,\cPtwo,\cPthree,\cPfour,\cPfive,\cPsix,\cPseven,\cPeight, $\cO$, $\cM^*$},
  ylabel = \rp,
  ylabel style={}, 
  every axis y label/.style={
    at={(ticklabel* cs:1.05)},
    anchor=south,
    font=\scriptsize},
  legend pos=south west,
  tick label style={font=\scriptsize},
  ymin=8,
  xmin=-2, xmax=13,
  minor y tick num=1,
  grid style={gray!1}, 
  height=4.5cm,
]
    \addplot[dino, mark=*, only marks] coordinates {(-1, 9.0)}; 
    \addplot[dinolineonly] coordinates {(-1.5, 9.0) (13, 9.0)};
    \addplot[pglevelsline] coordinates {
(1,	8.9)
(2,	9.1)
(3,	9.6)
(4,	9.7)
(5,	9.8)
(6,	9.8)
(7,	9.4)
(8,	7.5)
}; 
    \addplot[oracle, mark=triangle*] coordinates {(10, 9.8)};
    \addplot[grappaavg] coordinates {(12,  9.79)};
    \addplot[grappapg, mark=o] coordinates {(12, 9.8)}; 
    \addplot[grappa, mark=*] coordinates {(12, 9.9)}; 
\end{axis}
\end{tikzpicture}

%% file: plots/per_ds_cub.tex
\begin{tikzpicture}
\begin{axis}[%
  width=1.1\linewidth,
  title=\cub,
  xtick = {-1,1,2,3,4,5,6,7,8,10,12},
  xticklabels = {$\cM$, \cPone,\cPtwo,\cPthree,\cPfour,\cPfive,\cPsix,\cPseven,\cPeight, $\cO$, $\cM^*$},
  ylabel = \rp,
  ylabel style={}, 
  every axis y label/.style={
    at={(ticklabel* cs:1.05)},
    anchor=south,
    font=\scriptsize},
  legend pos=south west,
  tick label style={font=\scriptsize},
  ymin=30,
  ymax=36.5,
  xmin=-2, xmax=13,
  minor y tick num=1,
  grid style={gray!1}, 
  height=4.5cm,
]
    \addplot[dino, mark=*, only marks] coordinates {(-1, 33.6)}; 
    \addplot[dinolineonly] coordinates {(-1.5, 33.6) (13, 33.6)};
    \addplot[pglevelsline] coordinates {
    (1, 33.4)
    (2, 34.9)
    (3, 35.3)
    (4, 35.2)
    (5, 33.5)
    (6, 31.4)
    (7, 26.5)
    (8, 14.8)}; 
    \addplot[oracle, mark=triangle*] coordinates {(10, 35.3)}; 
    \addplot[grappaavg] coordinates {(12,  34.1)};
    \addplot[grappapg, mark=o, color=\grappacolor] coordinates {(12, 34.4)}; 
    \addplot[grappa, mark=*] coordinates {(12, 35.1)}; 
    
\end{axis}
\end{tikzpicture}

%% file: plots/per_ds_flowers.tex
\begin{tikzpicture}
\begin{axis}[%
  width=1.1\linewidth,
  title=\flowers,
  xtick = {-1,1,2,3,4,5,6,7,8,10,12},
  xticklabels = {$\cM$, \cPone,\cPtwo,\cPthree,\cPfour,\cPfive,\cPsix,\cPseven,\cPeight, $\cO$, $\cM^*$},
  ylabel = \rp,
  ylabel style={}, 
  every axis y label/.style={
    at={(ticklabel* cs:1.05)},
    anchor=south,
    font=\scriptsize},
  legend pos=south west,
  tick label style={font=\scriptsize},
  ymin=60,
  xmin=-2, xmax=13,
  minor y tick num=1,
  grid style={gray!1}, 
  height=4.5cm,
]
    \addplot[dino, mark=*, only marks] coordinates {(-1, 62.5)}; 
    \addplot[dinolineonly] coordinates {(-1.5, 62.5) (12, 62.5)};
    \addplot[pglevelsline] coordinates {
    (1,	63.3)
    (2,	66.7)
    (3,	67.9)
    (4,	67.9)
    (5,	67.5)
    (6,	66.9)
    (7,	64.8)
    (8,	56.7)
}; 
    \addplot[oracle, mark=triangle*] coordinates {(10, 67.9)}; 
    \addplot[grappaavg] coordinates {(12,  66.47)};
    \addplot[grappapg, mark=o] coordinates {(12, 66.53)}; 
    \addplot[grappa, mark=*] coordinates {(12, 67.2)}; 
\end{axis}
\end{tikzpicture}

%% file: plots/per_ds_food.tex
\begin{tikzpicture}
\begin{axis}[%
  width=1.1\linewidth,
  title=\food,
  xtick = {-1,1,2,3,4,5,6,7,8,10,12},
  xticklabels = {$\cM$, \cPone,\cPtwo,\cPthree,\cPfour,\cPfive,\cPsix,\cPseven,\cPeight, $\cO$, $\cM^*$},
  ylabel = \rp,
  ylabel style={}, 
  every axis y label/.style={
    at={(ticklabel* cs:1.05)},
    anchor=south,
    font=\scriptsize},
  legend pos=south west,
  tick label style={font=\scriptsize},
  ymin=25,
  ymax=31,
  xmin=-2, xmax=13,
  minor y tick num=1,
  grid style={gray!1}, 
  height=4.5cm,
]
    \addplot[dino, mark=*, only marks] coordinates {(-1, 27.0)}; 
    \addplot[dinolineonly] coordinates {(-1.5, 27.0) (13, 27.0)};
    \addplot[pglevelsline] coordinates {
(1,	29.0)
(2,	29.3)
(3,	29.5)
(4,	29.6)
(5,	29.5)
(6,	29.4)
(7,	28.8)
(8,	22.1)
}; 
    
    \addplot[oracle, mark=triangle*] coordinates {(10, 29.6)}; 
    \addplot[grappaavg] coordinates {(12,  30.37)};
    \addplot[grappapg, mark=o] coordinates {(12, 30.37)}; 
    \addplot[grappa, mark=*] coordinates {(12, 30.5)}; 
\end{axis}
\end{tikzpicture}

%% file: plots/per_ds_sop.tex
\begin{tikzpicture}
\begin{axis}[%
  width=1.1\linewidth,
  title=\sop,
  xtick = {-1,1,2,3,4,5,6,7,8,10,12},
  xticklabels = {$\cM$, \cPone,\cPtwo,\cPthree,\cPfour,\cPfive,\cPsix,\cPseven,\cPeight, $\cO$, $\cM^*$},
  ylabel = \rp,
  ylabel style={}, 
  every axis y label/.style={
    at={(ticklabel* cs:1.05)},
    anchor=south,
    font=\scriptsize},
  legend pos=south west,
  tick label style={font=\scriptsize},
  ymin=32.5,
  ymax=48,
  xmin=-2, xmax=13,
  minor y tick num=1,
  grid style={gray!1}, 
  height=4.5cm,
]
    \addplot[dino, mark=*, only marks] coordinates {(-1, 34.1)}; 
    \addplot[dinolineonly] coordinates {(-1.5, 34.1) (13, 34.1)};
    \addplot[pglevelsline] coordinates {
(1,	32.2)
(2,	32.7)
(3,	34.5)
(4,	35.8)
(5,	37.7)
(6,	39.7)
(7,	42.4)
(8,	46.5)
}; 
    
    \addplot[oracle, mark=triangle*] coordinates {(10, 46.5)}; 
    \addplot[grappaavg] coordinates {(12,  37.1)};
    \addplot[grappapg, mark=o] coordinates {(12, 35.9)}; 
    \addplot[grappa, mark=*] coordinates {(12, 36.3)}; 
\end{axis}
\end{tikzpicture}

%% file: tex/05_conclusions.tex
\section{Conclusions}
\label{sec:conclusions}

We present \grappa, an unsupervised approach for adapting a large pretrained backbone to simultaneously tackle multiple retrieval tasks, given only an unlabeled set of training images associated to these retrieval tasks. 
We show that one can adapt a large pretrained visual transformer using a set of pseudo-granularity adaptors and simple fusion layers. Our \grappa models bring consistent gains over the strong DINO~\cite{caron2021dino} baseline on all six retrieval tasks we adapt to. 
We envision this work as a first step towards models that dynamically adapt.

\myparagraph{Acknowledgements}
MIAI@Grenoble Alpes (ANR-19-P3IA-0003)

%% file: tex/99_appendix.tex
\section*{\Large Appendix}

We report further analysis that supports our choice to learn adaptors instead of fine-tuning the backbone (Section~\ref{sec:finetuning}), zero-shot performance of \grappa on datasets that are not in the \mrtlong benchmark (Section~\ref{sec:zeroshot}), and results on learning \grappa's fusion layer using class labels (Section~\ref{sec:supervised_fusion}).
We also present PyTorch-style pseudo-code for our architecture (Section~\ref{sec:pseudocode}).

\section{Learning adaptors vs fine-tuning the backbone}
\label{sec:finetuning}

We validate the different configurations of adaptors presented in the main paper and compare them in a supervised setting with a typical baseline that involves fine-tuning all layers of a given pretrained model $\cM$. In particular, we compare this baseline with fine-tuning just a single, randomly initialized adaptor ($\cM + \cA_1$), and with adding 8 random adaptors and fine-tuning them together with a fusion layer ($\cM^*$).

For both settings, the underlying model $\cM$ is \emph{kept fixed}, and we use DINO~\cite{caron2021dino} pretrained on ImageNet in all models.
We use the train splits of the 6 datasets in \mrt (\aircraft, \cars, \cub, \flowers, \food, and \sop) and train 6 individual models per method, \textit{using class labels} and a norm-softmax loss. 

We report the performance on their respective test splits in the diagonal of Table~\ref{tab:domain_shift_adaptors}. 
We observe that only adding a single adaptor outperforms the other models in almost all datasets. This is probably due to the fact that these are relatively small datasets and a single adaptor provides enough capacity to specialize, yet too little to overfit.
When comparing $\cM + \cA_1$ to $\cM^*$, we already saw a similar behavior in Table 2 in the main paper (rows 2 and 3), suggesting that randomly initialized adaptors 
perform worse when they are simultaneously trained  with a fusion layer, and that one way to address this is pretraining them for different objectives.

\section{Zero-shot performance } 
\label{sec:zeroshot}

\myparagraph{Supervised training separately on each dataset}
We consider specialized models that have been trained in a supervised manner on each dataset separately, \ie only on  one of the six train splits that compose \mrt. Our goal is to measure how much the performance of these specialized models drops when exposed to other retrieval tasks.
For this, we evaluate each of these models on the remaining five datasets and report their performance in the off-diagonal part of Table~\ref{tab:domain_shift_adaptors}. We can see how \emph{their performance drops significantly} compared to the pretrained DINO model, showing that they are overly suited to the retrieval task they have been trained on, and lose the generalization capability of the original pretrained model. It is also interesting to see that $\cM + \cA_1$ reports the lowest drops across all datasets, probably due to the fact that the underlying models are kept fixed and the adaptor's features are added in a residual fashion to the features of the backbone, retaining some of the generalization ability of the pretrained DINO model.

\input{tables/tab_domain_shift_with_adaptors}

\myparagraph{Supervised training jointly on all datasets}
Finally, we measure how much the zero-shot ability of the underlying model is altered after adapting it with \grappa on multiple datasets (\eg \mrt) using label supervision. We compare with other adaptation strategies: supervised fine-tuning of all layers, supervised fine-tuning of a single adaptor, and supervised fine-tuning of 8 adaptors and a fusion layer. For this experiment, we use three new datasets that were not part of \mrt: DTD~\cite{cimpoi2014texture}, Eurosat~\cite{helber2019eurosat}, and Pets~\cite{parkhi2012pets}.
We report \rp and mAP in Table~\ref{tab:zero_shot}.
First, we can see how little the zero-shot performance of \grappa drops compared to DINO, performing almost on par with DINO on DTD, and with only slightly larger drops on the other two datasets. Second, we can also see that compared to other supervised adaptation methods, retains a similar generalization ability.

\input{tables/tab_zero_shot}

\section{Supervised learning for adaptor fusion}
\label{sec:supervised_fusion}

We report additional results where the fusion layer over pseudo-granularity adaptors is learned in a supervised way. More precisely, for this experiment, we freeze adaptors \cPone-\cPeight and we learn the fusion layer on the training set of \mrt using class labels, a normsoftmax loss, and adam optimizer.
We report results on Table~\ref{tab:supervised_fusion}. As expected, we observe this fusion to significantly improve over \grappan in most of the datasets (we again stress that \grappa does not use any kind of labels). The difference is especially large on \sop, mostly likely due to imbalance issues; this dataset accounts for 97\% of the total number of classes.

\input{tables/tab_supervised_fusion}

\section{Pseudocode for a \grappa layer}
\label{sec:pseudocode}
In Algorithm~\ref{alg:grappa_layer} we show pseudocode for a layer of the \grappa model, consisting on a ViT block, followed by a set of adaptors, and a fusion layer combining the output of these adaptors. Details of this layer are also shown in Figure 3 in the main paper.

\begin{algorithm}[tb]
   \caption{PyTorch-style pseudocode for a \grappa layer.}
   \label{alg:grappa_layer}
   
    \definecolor{codeblue}{rgb}{0.5,0,0.5}
    \lstset{
      basicstyle=\fontsize{7.2pt}{7.2pt}\ttfamily\bfseries,
      commentstyle=\fontsize{7.2pt}{7.2pt}\color{codeblue},
      keywordstyle=\fontsize{7.2pt}{7.2pt},
    }
\begin{lstlisting}[language=python]
# B: batch size
# T: number of tokens
# D: dimensionality of the embeddings
# N: number of adaptors
#
# mm: matrix-matrix multiplication
# pool: average pooling
# MSA: multi-headed self attention
# MLP: multi-layer perceptron
# Adaptors: list of N bottleneck layers
# Q: learnable "query" DxD projection
# K: learnable "key" DxD projection
#
# h_1: input of size BxTxD

# ViT layer
x = MSA(LayerNorm(h_1)) + h_1
y = MLP(LayerNorm(x))
h = y + x  # BxTxD

# Compute N adaptors
U = []
for Adaptor in Adaptors:
    U.append(Adaptor(h) + y)
U = stack(U).permute(1, 2, 0, 3)  # BxTxNxD

# Fusion
q = mm(Q.T, pool(h, dim=1))  # BxD
k = mm(K.T, pool(U, dim=1))  # BxNxD
# Dot product between "query" and "key" to get the raw attention score
attn = mm(k.T, q.unsqueeze(-2))  # BxN
# Normalize the attention scores to probabilities
attn = softmax(attn, dim=1)
# Fuse adaptors using attention probabilities
f = mm(U.T, attn.unsqueeze(-2))  # BxTxD

h = f + x  # BxTxD

\end{lstlisting}
\end{algorithm}

%% file: tables/tab_domain_shift_with_adaptors.tex
\begin{table}[t]
    \centering
    \caption{
    \textbf{\textcolor{red}{Supervised} finetuning using \textcolor{red}{class and task labels}}, applied to the current task and to other retrieval tasks.
    Based on the ImageNet pre-trained DINO model $\cM$, each model is finetuned on a single dataset in a supervised manner (row) and tested on the same and on the other datasets (column). We report \mapatr (mAP) and \rp. Note that those numbers constitute an \textbf{upperbound} and they are \textcolor{red}{not} directly comparable with those from the main paper.}
    \label{tab:domain_shift_adaptors}
    \setlength{\tabcolsep}{5pt}
    \resizebox{\linewidth}{!}{
    \begin{tabu}{l  rr rr rr rr rr rr}
    \toprule
      \multirow{2}{*}{Train \textbackslash~Test} & \multicolumn{2}{c}{~~~~~~\aircraft} & \multicolumn{2}{c}{~~~~~~\cars} & \multicolumn{2}{c}{~~~~~~\cub} & \multicolumn{2}{c}{~~~~~~\flowers} & \multicolumn{2}{c}{~~~~~~\food} & \multicolumn{2}{c}{~~~~~~\sop} \\
         & ~~~~~~~~\rp         & mAP      & ~~~~~~~~\rp           & mAP        & ~~~~~~~~\rp         & mAP      & ~~~~~~~~\rp           & mAP        & ~~~~~~~~\rp           & mAP         & ~~~~~~~~\rp           & mAP        \\
        
        \midrule

        DINO ($\cM$) & 17.5 & 9.2 & 9.0 & 3.5 & 33.6 & 22.9 & 62.5 & 57.0 & 27.0 & 16.1 & 34.1 & 31.5 \\ 
        \midrule      
        \multicolumn{13}{l}{~\textbf{\emph{\aircraft}}}  \vspace{3pt} \\ 
        $\cM$ (supervised finetuning) & \textbf{33.3} & \textbf{20.5} & 7.7 & 2.7 & 17.2 & 8.6 & 56.2 & 49.5 & 14.3 & 5.5 & 33.9 & 31.3  \\
        $\cM$ + $\cA_1$ (supervised) & \textbf{33.2} & \textbf{20.9} & 10.5 & 4.4 & 23.2 & 13.4 & 59.4 & 53.1 & 18.3 & 8.6 & 30.0 & 27.5\\ 
        $\cM^*$ (random, supervised) & 19.9 & 10.1 & 8.4 & 3.0 & 12.1 & 5.1 & 54.9 & 47.6 & 13.8 & 5.2 & 29.0 & 26.6  \\
        \midrule
        
        \multicolumn{13}{l}{~\textbf{\emph{\cars}}}  \vspace{3pt} \\ 
        $\cM$ (supervised finetuning) & 15.9 &  8.0 &  \textbf{35.2} &  \textbf{24.9} &  18.6 &  9.7 &  59.0 &  52.6 &  14.0 &  5.4 &  30.2 &  27.7 \\
        $\cM$ + $\cA_1$ (supervised)  & 19.6 &  10.9 &  33.4 &  23.1 &  28.2 &  18.0 &  62.3 &  56.4 &  19.3 &  9.4 &  28.7 &  26.2 \\ 
        $\cM^*$ (random, supervised)  & 12.2 & 5.1 & 8.3 & 2.7 & 10.8 & 4.2 & 49.1 & 41.1 & 11.7 & 3.9 & 26.4 & 24.2  \\
        \midrule
        
        \multicolumn{13}{l}{~\textbf{\emph{\cub}}}  \vspace{3pt} \\ 
        $\cM$ (supervised finetuning)  & 14.1 &  6.4 &  7.5 &  2.7 &  39.1 &  28.2 &  50.0 &  42.7 &  16.0 &  6.7 &  34.8 &  32.1 \\
        $\cM$ + $\cA_1$ (supervised)  & 16.4 &  8.4 &  9.2 &  3.7 &  \textbf{45.0} &  \textbf{34.8} &  58.7 &  52.9 &  23.2 &  12.6 &  36.5 &  33.7 \\ 
        $\cM^*$ (random, supervised) & 16.4 & 8.4 & 9.1 & 3.6 & 42.0 & 31.5 & 59.2 & 53.4 & 24.4 & 13.7 & 34.5 & 31.8  \\
        \midrule

        \multicolumn{13}{l}{~\textbf{\emph{\flowers}}}  \vspace{3pt} \\ 
        $\cM$ (supervised finetuning) & 14.7 &  6.7 &  7.3 &  2.5 &  21.1 &  11.6 &  68.7 &  63.5 &  15.9 &  6.7 &  34.5 &  31.8 \\
        $\cM$ + $\cA_1$ (supervised)  & 17.3 &  8.9 &  9.1 &  3.5 &  29.9 &  19.3 &  \textbf{73.8} &  \textbf{69.6} &  23.0 &  12.5 &  36.1 &  33.4 \\ 
        $\cM^*$ (random, supervised) & 17.2 & 8.7 & 9.6 & 3.7 & 28.3 & 17.9 & 68.1 & 62.6 & 20.0 & 9.9 & 32.3 & 29.8  \\
        \midrule
        
        \multicolumn{13}{l}{~\textbf{\emph{\food}}}  \vspace{3pt} \\ 
        $\cM$ (supervised finetuning) & 8.4 &  2.8 &  5.0 &  1.4 &  8.7 &  2.9 &  44.0 &  35.8 &  30.9 &  18.8 &  25.0 &  22.8 \\
        $\cM$ + $\cA_1$ (supervised)  & 16.7 &  8.6 &  8.5 &  3.1 &  35.1 &  24.3 &  56.0 &  49.4 &  \textbf{36.2} &  \textbf{24.9} &  35.2 &  32.5 \\ 
        $\cM^*$ (random, supervised) & 17.5 & 9.3 & 8.9 & 3.4 & 35.6 & 24.8 & 60.3 & 54.4 & 34.1 & 22.9 & 35.5 & 32.8 \\
        \midrule
    
        \multicolumn{13}{l}{~\textbf{\emph{\sop}}}  \vspace{3pt} \\ 
        $\cM$ (supervised finetuning) & 7.2 &  2.3 &  4.9 &  1.6 &  5.6 &  1.5 &  36.6 &  27.9 &  8.8 &  2.2 &  42.4 &  39.7 \\
        $\cM$ + $\cA_1$ (supervised) & 12.1 &  5.2 &  7.0 &  2.4 &  20.8 &  11.2 &  53.0 &  46.0 &  15.0 &  6.1 &  \textbf{53.6} &  \textbf{50.9} \\ 
        $\cM^*$ (random, supervised) & 13.3 & 5.9 & 6.9 & 2.3 & 24.0 & 13.7 & 54.9 & 48.1 & 18.9 & 9.1 & 49.0 & 46.2 \\

    \bottomrule
    \end{tabu}
    }
    
\end{table}

%% file: tables/tab_zero_shot.tex
\begin{table}[t]
    \centering
    \caption{\textbf{Performance of models trained on \mrt, and tested on other retrieval tasks}. We measure the drop in generalization (zero-shot performance) after adapting the DINO model with \grappa versus alternative ways. Training is always performed on \mrt, in a supervised manner for all methods (middle section), except for \grappa which is fully unsupervised (bottom section). We report \mapatr (mAP) and \rp.}
    \label{tab:zero_shot}
    \setlength{\tabcolsep}{5pt}
    \resizebox{\linewidth}{!}{
    \begin{tabu}{l c rr rr rr}
    \toprule
      \multirow{2}{*}{Train \textbackslash~Test} & Supervised? & \multicolumn{2}{c}{~~~DTD} & \multicolumn{2}{c}{~~~Eurosat} & \multicolumn{2}{c}{~~~Pets}\\
           &  & ~~~~~\rp         & mAP      & ~~~~~\rp           & mAP        & ~~~~~\rp         & mAP   \\
        
        \midrule
        $\cM$ (DINO -- zero-shot model) & - & 41.5 & 30.6 & 76.3 & 52.5 & 76.0 & 70.9 \\ \midrule
        $\cM$ (supervised finetuning) & \cmark & 23.1 & 12.7 & 65.8 & 41.9 & 23.4 & 10.5\\
        $\cM$ + $\cA_1$ (supervised) & \cmark & 39.2 & 28.5 & 76.6& 52.8 & 74.1 & 68.7  \\
        $\cM^*$ (random, supervised) & \cmark & 39.5 & 28.7 & 75.8 & 51.7 & 73.2 & 67.5  \\
        \midrule
        \grappan & \xmark & 41.3 & 30.5 & 75.2 & 51.0 & 74.4 & 68.9 \\
        ~~\emph{drop versus DINO} & & \diffdown{0.2} & \diffdown{0.1} & \diffdown{1.1} & \diffdown{1.5} & \diffdown{1.6} & \diffdown{2.0} \\
        ~~\emph{gains over best adapted} & & \diffup{1.8} & \diffup{1.8} & \diffdown{1.4} & \diffdown{1.8} & \diffup{0.3} & \diffup{0.2}\\
    \bottomrule
    \end{tabu}
    }
    
\end{table}

%% file: tables/tab_supervised_fusion.tex
\begin{table}[t]
    \centering
    \caption{
    \textbf{Performance of a model trained on MRT, using supervised learning of the fusion layer {\color{red}{using class labels}} (supervised fusion).}  We report \mapatr (mAP) and \rp on the six tasks of the \mrt benchmark. We compare with \grappan which is unsupervised.}
    \label{tab:supervised_fusion}
    \setlength{\tabcolsep}{5pt}
    \resizebox{\linewidth}{!}{
    \begin{tabu}{l   rr rr rr rr rr rr}
    \toprule
     & \multicolumn{2}{c}{~~~~~~\aircraft} & \multicolumn{2}{c}{~~~~~~\cars} & \multicolumn{2}{c}{~~~~~~\cub} & \multicolumn{2}{c}{~~~~~~\flowers} & \multicolumn{2}{c}{~~~~~~\food} & \multicolumn{2}{c}{~~~~~~\sop} \\
            & ~~~~~~~~\rp         & mAP      & ~~~~~~~~\rp           & mAP        & ~~~~~~~~\rp         & mAP      & ~~~~~~~~\rp           & mAP        & ~~~~~~~~\rp           & mAP         & ~~~~~~~~\rp           & mAP        \\
        
        \midrule
\grappan  & 18.1 & 9.5 & 9.9 & 4.0 & 35.1 & 24.1 & 67.2 & 61.9 & 30.5 & 19.3 & 36.3 & 33.6 \\
Supervised fusion & 19.8 & 10.7 & 11.1 & 4.7 & 34.7 & 23.7 & 67.6 & 62.2 & 30.8 & 19.5 & 43.2 & 40.3 \\
    \bottomrule
    \end{tabu}
    }
    
\end{table}